%% file: main.tex
\documentclass[journal]{IEEEtai}

\usepackage[colorlinks,urlcolor=blue,linkcolor=blue,citecolor=blue]{hyperref}
\usepackage{svg} 
\newcommand{\orcidicon}[1]{%
    \href{https://orcid.org/#1}{\includegraphics[width=1.8ex]{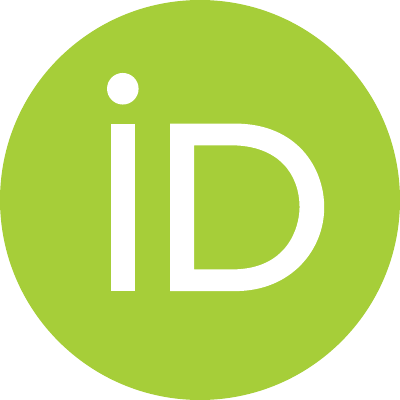}}%
}

\usepackage{color,array}

\usepackage{graphicx}
\usepackage{algorithm}
\usepackage{algpseudocode}
\usepackage{array}
\usepackage{amsmath,amsfonts}
\usepackage[caption=false,font=normalsize,labelfont=sf,textfont=sf]{subfig}
\usepackage{textcomp}
\usepackage{stfloats}
\usepackage{url}
\usepackage{verbatim}
\usepackage{cite}

\usepackage[justification=centering]{caption}
\usepackage{algpseudocode}
\algblock{Input}{EndInput}
\algnotext{EndInput}
\algblock{Paramters}{EndParameter}
\algnotext{EndParameter}
\newcommand{\Desc}[2]{\State \makebox[2em][l]{#1}#2}
\usepackage{multicol}
\usepackage{multirow}
\usepackage{adjustbox}
\usepackage{comment}
\usepackage{xcolor}
\usepackage[numbers]{natbib}

\newcommand{\saransh}[1]{\textcolor{black}{#1}}
\begin{document}

\title{IDALC: A Semi-Supervised Framework for Intent Detection and Active Learning based Correction} 

\author{Ankan Mullick\orcidicon{0000-0002-0721-1359}, Sukannya Purkayastha\orcidicon{0000-0002-7559-0522}, Saransh Sharma\orcidicon{0009-0001-7662-3496}, Pawan Goyal\orcidicon{0000-0002-9414-8166}, and Niloy Ganguly\orcidicon{0000-0002-3967-186X}, \IEEEmembership{Senior Member, IEEE}

\thanks{Ankan Mullick, Pawan Goyal, and Niloy Ganguly are with the Department of Computer Science and Engineering, IIT Kharagpur, India (e-mail: ankanm@kgpian.iitkgp.ac.in).}
\thanks{Sukannya Purkayastha is with the Department of Computer Science, Technische Universität Darmstadt, Germany. Work done while at IIT Kharagpur.}
\thanks{Saransh Sharma is with Adobe Research, Bangalore, India (e-mail: sarsharma@adobe.com).}}

\markboth{Journal of IEEE Transactions on Artificial Intelligence}
{Mullick \MakeLowercase{\textit{et al.}}: IDALC: A Semi-Supervised Framework for Intent Detection and Active Learning based Correction}
\maketitle

\begin{abstract}
Voice-controlled dialog systems have become immensely popular due to their ability to perform a wide range of actions in response to diverse user queries. These agents possess a predefined set of skills or intents to fulfill specific user tasks. But every system has its own limitations. There are instances where even for known intents, if any model exhibits low confidence, it results in rejection of utterances that necessitate manual annotation. Additionally, as time progresses, there may be a need to re-train these agents with new intents from the system-rejected queries to carry out additional tasks. Labeling all these emerging intents and rejected utterances over time is impractical, thus calling for an efficient mechanism to reduce annotation costs. In this paper, we introduce IDALC (\textbf{I}ntent \textbf{D}etection and \textbf{A}ctive \textbf{L}earning based \textbf{C}orrection), a semi-supervised framework designed to detect user intents and rectify system-rejected utterances while minimizing the need for human annotation. Empirical findings on various benchmark datasets demonstrate that our system surpasses baseline methods, achieving a $\sim$5-10\% higher accuracy and a $\sim$4-8\% improvement in macro-F1. Remarkably, we maintain the overall annotation cost at just $\sim$6-10\% of the unlabelled data available to the system. {The overall framework of IDALC is shown in Fig.~\ref{fig:idalc_overview}.}

\end{abstract}

\begin{IEEEImpStatement}
Conversational AI systems are widely used across industries, yet they often reject user inputs due to low confidence in intent recognition. This results in poor user experience and high manual annotation costs. Our framework, IDALC, offers a practical alternative by combining intent detection with active learning to automatically correct low-confidence predictions. It reduces labeling effort by over 90\% while maintaining high accuracy across languages and domains. Unlike large language models, which are resource-intensive and difficult to deploy in real-time or on edge devices, IDALC is lightweight, efficient, and more adaptable to evolving user needs. It performs comparably, or even better, than LLMs for this task, without the heavy computational burden. By making intent detection more scalable and accessible, IDALC has strong potential to support voice-driven services in education, healthcare, and public-facing digital assistants where speed, cost, and reliability matter most.
\end{IEEEImpStatement}


\input{1Intro}
\input{2Related}

\input{3Dataset}
\input{4Approach}
\input{5Experiment}

\input{6Error}

\section{Limitations}

\saransh{While our study shows promising results, it has several limitations that open important directions for future work: (1) ID-ALC relies on human-in-the-loop feedback, which improves accuracy but requires costly manual annotation; we plan to explore zero-shot adaptation using self-learning and large language models. (2) The framework currently supports only text, whereas real-world applications often involve voice, images, or mixed modalities; we will scale datasets and replace classifiers with multimodal models such as SDIF-DA~\cite{huang2023sdifda}, MulT~\cite{tsai-etal-2019-multimodal}, MISA~\cite{hazarika2020misa}, and MAG-BERT~\cite{rahman-etal-2020-integrating}. (3) Our experiments are limited to English, Thai, and Spanish, leaving out code-mixed or code-switched settings; extending to such cases will require multilingual models with alignment across framework components. (4) We also plan to evaluate low-resource languages and domain-specific datasets (e.g., education, healthcare, law) to test adaptability beyond general-purpose intent detection. (5) While the current work focuses on optimizing performance with minimal annotation, future extensions will address the interpretability of IDALC, as understanding which features or interactions drive clustering and adaptation can improve trust, usability, and applicability in expert-facing domains.}

\section{Conclusion}

\saransh{We present \textbf{IDALC} (Intent Detection and Active Learning based Correction), a semi-supervised framework designed to handle rejected utterances and discover new intents with minimal human annotation. IDALC combines active learning with majority voting to select informative samples, enabling iterative improvement while reducing annotation cost. The framework is scalable, integrates easily with existing dialog systems, and remains effective even in low-resource scenarios. Experiments on multiple benchmarks demonstrate consistent gains over baselines, with 5–10\% higher accuracy and 4–8\% better macro-F1, while requiring annotation for only 6–10\% of unlabeled data. Looking ahead, we aim to extend IDALC to dynamically evolving intent classes, incorporate zero- and few-shot techniques for further reducing supervision, and study interpretability to understand which features and interactions drive model decisions. We also plan to evaluate the framework in real-world domains such as finance, healthcare, and law, where both efficiency and transparency are critical. IDALC can serve as a building block for next-generation dialog systems with structured intent understanding, ensuring reliable, adaptive, and trustworthy user interactions.}

\section*{Ethical Implications}
We use publicly available datasets and do not contain any sensitive personal information or harmful content. They come with existing annotations, which we use as labels. During prediction, we mask these labels and predict them. We also have not used any AI-generated text.

\bibliographystyle{IEEEtran}
\bibliography{sample-base}

\end{document}

%% file: 1Intro.tex
\section{Introduction}

\IEEEPARstart{O}ver the past few years, task-oriented dialog systems have become ubiquitous. The use of these agents to perform tasks based on voice commands has opened up a new dimension in the field of Natural Language Understanding (NLU). These agents are characterized by various skills to perform various tasks based on a user's query. These predefined sets of skills are known as intents. For eg., the query \textbf{`add Michael Wittig music to country icon playlist'} requires to do the task of adding `Michael Wittig music' to the `country icon playlist' and it would be associated with the \textit{AddToPlaylist} intent category (SNIPS Dataset). The task-oriented systems have some initial set of known intents that can easily be identified and responded to accordingly, but 
when users interact with voice-controlled systems or chatbots, there are instances where the system fails to comprehend the user's query or classify it wrongly into a known intent. 
The system even struggles to classify complex utterances from the known set of intents due to various reasons such as ambiguity, uncommon user requests, or untrained intents. These hard-instanced queries eventually get rejected because of lower confidence. Effective handling of system-rejected search queries is a critical challenge. To improve the user experience and enhance the system's performance, resolving this issue by understanding novel intents and detecting known intents with higher accuracy is essential.

\begin{figure}
	\centering
    	\fbox{\includegraphics[width=0.8\linewidth]{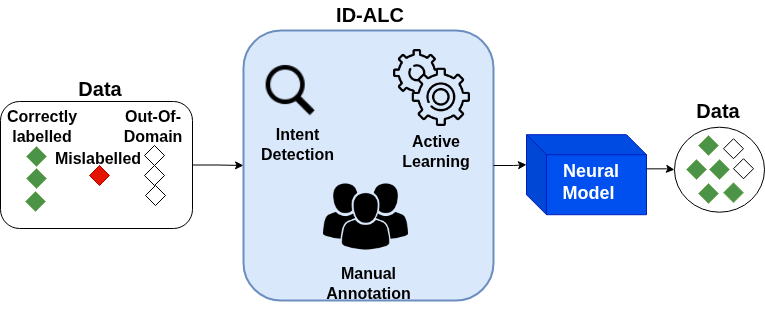}}
	\caption{Overview of IDALC Framework:  The system processes \\ user utterances, identifies known intents with high confidence, \\ and flags low-confidence or unknown ones for correction}
    \vspace{-5mm}
	\label{fig:idalc_overview}
\end{figure}

\begin{figure*}
    \centering
    \includegraphics[width=16cm, height=6cm]{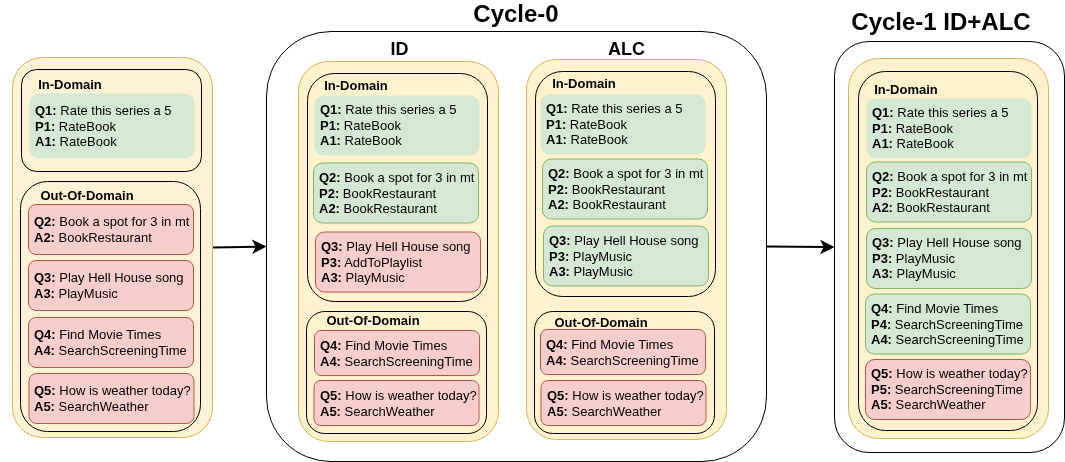}
    \caption{End-to-end framework of IDALC (Intent Detection and Active Learning-based Correction) is illustrated using predicted (\textbf{P1–P5}) and actual intent labels (\textbf{A1–A5}). Initially, only two intents, RateBook and AddToPlayList, are known. Q1 is correctly classified. Q2–Q5 have unknown intents. In Cycle 0, Q2 is correctly classified during the Intent Detection (ID) step. Q3 is incorrectly predicted as 'AddToPlayList' but belongs to a new intent: 'PlayMusic'. As the system is unsure, it rejects Q3. In the ALC step, Q3 is corrected, and PlayMusic is added to the list of known intents. In the next cycle, Q4 is correctly classified as SearchScreeningTime. A5 is added to the set of known intents, but Q5 is still misclassified. After each ID step, some previously unknown queries are mapped to known intents. After each ALC step, incorrect predictions are corrected. By the end, some queries may remain misclassified or unrecognized.}
    \vspace{-4mm}
    \label{fig:idalc_archi_final}
    \end{figure*}


To address the challenge of handling system-rejected queries, several requirements must be met. First, a comprehensive evaluation of multiple datasets needs to be set up to analyze the problem of rejected queries and the necessity for manual annotation. 
Moreover, the development of a robust intent detection algorithm plays a crucial role in accurately classifying queries into their respective intents. So, with the emerging new intents in the system, the conversational agents need to be retrained to identify these newer intents aka \textit{novel classes} without compromising known intent accuracy. 
However, it is infeasible to label all the new instances as well as the rejected utterances because of the sheer volume, and hence an effective mechanism to reduce human annotation cost would be required.  
So, an algorithm should be designed to handle both known and novel intents, ensuring that the system can adapt and recognize emerging user requests effectively with optimized annotation. Active learning methods can further enhance the intent detection model by selecting the most informative queries for manual annotation. Regular retraining of the model with new data, including rejected queries and annotated queries, allows the system to continually improve its performance and stay up-to-date with evolving user intents.

In this paper, we adopt a two-step semi-supervised setting to identify known as well as novel out-of-domain (OOD) intents using an active learning framework. We also experiment with different scenarios of similarities among known and novel intents. Corresponding to our problem setting, we start with an initial labeled training data which consists of a set of known intents and a large unlabeled data that comprises known and novel intents. The evaluation is performed on a predefined test set that consists of all the known and new classes. Our framework consists of two architectures running sequentially - Intent Detection (ID) and Active Learning based Correction (ALC). Fig.~\ref{fig:idalc_overview} gives a high-level view of the IDALC framework. It shows how the system processes user queries by first identifying known intents with high confidence. If the system is unsure or encounters a new intent, it flags the query for correction through active learning. 

Fig.~\ref{fig:idalc_archi_final} provides a detailed step-by-step example of how this process works. At the start, the system only knows two intents, RateBook and AddToPlayList. It correctly identifies the intent of the first query (Q1), but the intents for Q2, Q3, and Q4 are unknown. In the first cycle (Cycle 0), the system processes Q2 and Q3. It gets Q2 right but misclassifies Q3 as AddToPlayList, a known intent, even though the correct label is PlayMusic. This mistake is caught and corrected in the Active Learning phase. In the next cycle (Cycle 1), the system has now learned the intent behind Q4 and can correctly classify it.

Our system is compared with similar models and evaluated across several NLU standard public datasets. 
Our framework is language agnostic and performs significantly well on different multilingual (English, Spanish, and Thai) datasets. IDALC outperforms the existing baselines by a margin of $\sim$ 4-8\% macro-F1 and $\sim$ 5-10\% in terms of accuracy on various datasets. We also show that our model is able to achieve better outcomes than large language models like ChatGPT, and LLaMA-2. The annotation cost incurred by our model varies from 6-10\% depending on the size of the unlabeled dataset. 

%% file: 2Related.tex
\section{Related Work}
In the last decade, researchers have focused on intent detection in various scenarios. People also explore various active learning methods to get optimum accuracy with minimum labeling cost. We review related work on intent class detection as well as active learning-based strategies to improve classification and lower human effort.

\subsection{Intent Class Detection}
In dynamic and evolutionary real-world applications, various flexible adaptive intent class detection models are developed. \cite{liao2023novel,kuzborskij2013n, scheirer2012toward, degirmenci2022efficient} work on incremental learning in a dynamic environment for evolving new classes in streams. \cite{sun2016online} aims at the class evolution phenomenon of the emergence and disappearance of classes. Some advanced approaches are proposed to explore novel class detection problems in data streams. ECSMiner \cite{masud2010classification} 
resolve streaming problems 
but the time delay assumption 
and fixed-size chunks are unrealistic in real scenarios. SAND \cite{haque2016sand} is a semi-supervised framework to identify concept drifts. 
\cite{mullick2022framework} identifies novel intents but does not apply any approach to retain the accuracy of known intents and does not handle system-rejected utterances. 
In different specialized contexts, researchers also work on the intent phrase identification \cite{mullick2024pointer,mullick2023intent,mullick2025text}
Based on the Frequent-Directions technique \cite{liberty2013simple}, SENC-MaS \cite{mu2017streaming} is a matrix sketch method to detect new classes and update the classifier. SENCForest \cite{mu2017classification} utilizes random trees as a single common core to identify new classes. But both these approaches rely on a large
labelled initial training set and do not use unlabelled data. 
\cite{cui2023novel,abdallah2016anynovel,zhou2022detecting} explore outlier detection approaches but most of them are applicable on images. \cite{na2018dilof,zhan2021out,larson2019evaluation,yan2020unknown, zhou2022knn, firdaus2023multitask} detect new class in the form of outlier detection. However, these algorithms fail to identify new classes from rejected utterances and classify all. SEEN \cite{zhu2020semi} 
uses both labeled and unlabelled data to detect emerging new classes and classify known classes. 
Zero-shot-OOD \cite{tan2019out} proposes
a new method to tackle the OOD detection task
in low-resource settings, without adversely affecting performance on the few-shot ID classification. 
Aligner \cite{zhu2024aligner2}, CEA-Net \cite{wu2024cea} explore joint multiple intent detection with slot filling task. \cite{park2024joint} apply the joint-BERT model for domain-specific intent classification.
SENNE \cite{cai2019nearest} and IFSTC \cite{xia2021incremental} handle streaming class classification problems. TARS \cite{halder-etal-2020-task} identifies in-domain data and classifies them. We use SEEN, SENC-MaS, Zero-shot-OOD, SENNE, IFSTC, and TARS as baselines. 

\subsection{Active Learning based Correction}
Active learning (AL) is a machine learning paradigm that aims to reduce the annotation effort by selectively choosing the most informative data points to label. Instead of randomly labeling large volumes of data, AL focuses on identifying uncertain or representative instances that are likely to improve the model's performance when labeled. This is particularly useful in tasks like intent detection, where obtaining labeled data can be expensive and time-consuming, especially when dealing with large or diverse datasets. \cite{wang2020active} proposes an active learning strategy, based on label error statistical methods to find critical instances. \cite{salama2022novel} uses active learning for heterogeneous data.   \cite{senay2020viraal, zhang2020deep} 
 apply active learning (AL) method for intent classification. \cite{yin2023clustering} is a cluster-based active learning method works on data streams.  
 But none of these works focused on accuracy improvements of intent classes using AL approaches. \cite{gao2020consistency} presents a consistency-based semi-supervised AL framework and a simple pool-based AL selection metric to select data for labeling. But these models do not handle system-rejected samples and are useful for images, not in textual contexts. \cite{modAL2018} proposes various AL strategies to improve system accuracy. \cite{tian2011active} develops a maximal marginal uncertainty-based AL model to maximize accuracy with minimal interactions. We use the last two approaches as baselines (MaxMU-AL and MinU-AL). 

 None of the earlier works focus on developing a single system to utilize labelled and unlabelled data together to detect unknown intents, improve known intent identification with optimized human annotation cost. Our approach can improve intent detection accuracy while handling system-rejected utterances and also lower the human labelling task.


%% file: 3Dataset.tex
\begin{table*}[!htb]
\vspace{-2mm}
\centering
\begin{adjustbox}{width=0.768\linewidth}
\begin{tabular}{|c|c|c|c|c|c|c|c|c|c|c|}
\hline
\multirow{2}{*}{\textbf{Dataset}} & \multirow{2}{*}{\textbf{\begin{tabular}[c]{@{}c@{}}\# Intent \\ Class\end{tabular}}}  & \multirow{2}{*}{\textbf{Total data}} & \multirow{2}{*}{\textbf{Type}} & \multirow{2}{*}{\textbf{\# Lab}} & \multicolumn{3}{c|}{\textbf{\#Unlab}}            & \multicolumn{3}{c|}{\textbf{\#Test}}             \\ \cline{6-11} 
                                  &                                               &                                    &                                &                                  & \textbf{\#Ind} & \textbf{\#OOD} & \textbf{\#Tot} & \textbf{\#Ind} & \textbf{\#OOD} & \textbf{\#Tot} \\ \hline
\multirow{3}{*}{SNIPS}            & \multirow{3}{*}{7 (5  + 2 )}                                                                                                                      & \multirow{3}{*}{14484}             & 1                              & 2990                             & 5661           & 3449           & 9110           & 1679           & 705            & 2384           \\ \cline{4-11} 
                                                                                                                       &                                                                                  &                                    & 2                              & 2990                             & 5629           & 3481           & 9110           & 1709           & 675            & 2384           \\ \cline{4-11} 
                                  &                                                                                                                                                                       &                                    & 3                              & 2990                             & 5689           & 3421           & 9110           & 1692           & 692            & 2384           \\ \hline
\multirow{3}{*}{FB}               & \multirow{3}{*}{12 (9 + 3)}                                                                                                                       & 43323                              & EN                             & 10000                            & 9978           & 15022          & 25000          & 3849           & 4474           & 8323           \\ \cline{3-11} 
                                                                                                                       &                                                                                  & 8643                               & ES                             & 2000                             & 2321           & 1678           & 3999           & 1512           & 1132           & 2644           \\ \cline{3-11} 
                                  &                                                                                                                                                                       & 5083                               & TH                             & 1000                             & 1203           & 1797           & 3000           & 318            & 765            & 1083           \\ \hline
ATIS                              & 17 (10  + 7 )                                                                                                                                                    & 5871                               & -                              & 1501                             & 2688           & 682            & 3370           & 802            & 198            & 1000           \\ \hline
\end{tabular}
\end{adjustbox}
\vspace{1mm}
\caption{
Dataset statistics based on our custom splits for multiple intent detection benchmarks. The datasets include SNIPS, Facebook Multilingual (FB), and ATIS. Each dataset is partitioned into labeled, unlabeled, and test subsets, further broken down into in-distribution (Ind) and out-of-distribution (OOD) samples. For SNIPS, we report three different Type settings.}
\vspace{-4mm}
\label{tab:dataset}
\end{table*}

\section{Dataset}

We perform our experiments on three NLU benchmarked datasets  - SNIPS \cite{couckesnips:2018}, ATIS \cite{tur2010left} and Facebook Multi-lingual data \cite{schuster2018cross}. 
   
 \noindent \textbf{(A) SNIPS}: This dataset is collected from the SNIPS personal voice assistant. It has a total of 14,484 sentences with 7 intent types. We perform our experiments considering 5 intents as known and the rest 2 as unknown / novel. The split of the dataset is shown in Table \ref{tab:dataset}. \textcolor{black}{We use acronyms such as \textbf{BR} (BookRestaurant), \textbf{AP} (AddToPlaylist), \textbf{GW} (GetWeather), \textbf{RB} (RateBook), \textbf{SSE} (SearchScreeningEvent), \textbf{SCW} (SearchCreativeWork) and \textbf{PM} (PlayMusic) to represent the intents.} To understand the effect of choosing various types of novel intents and how the system performs in different scenarios, we perform 3 different types of division as follows:
    
(i) \textbf{Type 1} (\textit{Unknowns are dissimilar from each other but similar to the known intents}): For this experiment, we consider `SCW' and `PM' as unknowns which resemble the known intents, `SSE' and `AP', respectively.
(ii) \textbf{Type 2} (\textit{Unknowns are dissimilar from each other and also from the known intents}): In this case, we consider `GW' and `RB' as unknown intents which are distinct classes and bear no resemblance to any other classes.
(iii) \textbf{Type 3} (\textit{Unknowns are similar to each other but dissimilar from the known intents}): The intents `SSE' and `SCW' are considered as unknown intents here which are similar to each other but have no similarity with any other known intents. 

\noindent \textbf{(B) Facebook-multilingual Dataset}: The dataset consists of data from three different languages: English (FB-EN), Spanish (FB-ES), and Thai (FB-TH) which consist of 43,323, 8,643, and 5,083 utterances, respectively. The utterances have been collected from three domains: Alarm, Reminder, and Weather with 12 different intents. We consider 9 intents as known and the rest 3 as unknown. 

\noindent \textbf{(C) ATIS Dataset}: Airline Travel Information System (ATIS) consists of audio recordings of people making flight reservations. It consists of 5,871 user queries with 17 different intent types. 
We consider 10 intents as known and 7 as unknown. 

The detailed statistics of these datasets, including our experimental framework are shown in Table \ref{tab:dataset}. We take the existing benchmarked NLU publicly available datasets, omit a few intents, mark them as OOD, and detect them. Since the datasets are fully labeled, annotation here just means using the given labels. We use them for training and hide them during prediction, so there are no issues like labeling errors or differences between annotators, and the data stays consistent. 

%% file: 4Approach.tex
\section{Approach}
We propose the algorithm ``ID-ALC'' (Intent Detection and Active Learning based Correction) which is shown in Algorithm \ref{algo:mnid-alc} with two major procedures: intent detection (ID - Algorithm \ref{algo:id}) and active learning based correction (ALC - Algorithm \ref{algo:alc}). The 
ID module detects out-of-domain samples on unlabeled data, identifies new intent classes as described in Sec \ref{sec:ood} and the 
ALC, described in Sec \ref{sec:alc}, handles the rejected utterances while reducing the annotation cost and also ensuring higher overall accuracy on the known intents. In our framework, we keep adding data to the initial pool of labeled data at every step using the two components (running multiple iterations), we call each of these model retraining steps as \textit{cycles}. The  ID-ALC framework is shown in Fig. \ref{fig:al_archi_final} with its two components.

\begin{algorithm}[!thb]
\caption{Intent Detection and Active Learning based Correction (ID-ALC) Algorithm}
\label{algo:mnid-alc}
\begin{algorithmic}[1]
\Input
  \Desc{$\mathcal{L}$}{\hspace{2mm}Labelled Data (\textbf{Cycle 0})}
  \Desc{$\mathcal{U}$}{\hspace{2mm}Unlabelled Data }
   \Desc{$\mathcal{T}$}{\hspace{2mm}Blind Test Data}
  \Desc{$\mathcal{TH}$}{\hspace{2mm}A predefined Threshold}
  \Desc{$\mathcal{MV}$}{\hspace{2mm}Majority Voting Count}
  \EndInput
\Paramters
    \Desc{$\mathcal{M}$}{Neural Model}
    \EndParameter
\State $\mathcal{L}$, $\mathcal{U}_{rem}$, $\mathcal{M} \gets $ ID($\mathcal{L}, \mathcal{U}, \mathcal{M}$, $\mathrm{T}$) 

\State $\mathcal{L}$, $\mathcal{U}_{rem}$, $\mathcal{M} \gets $ ALC($\mathcal{L}, \mathcal{U}_{rem}, \mathcal{M}$, $\mathrm{T}$, $\mathcal{TH}$, $\mathcal{MV}$) 
\end{algorithmic}
\end{algorithm}

\vspace{-2mm}

\begin{algorithm}[!thb]
\caption{Intent Detection: ID($\mathcal{L}, \mathcal{U}, \mathcal{M}$, $\mathrm{T}$)}
\label{algo:id}
\begin{algorithmic}[1]
\Procedure{Intent Detection}{}
\State Train $\mathcal{M}$ on $\mathcal{L}$ and predict on $\mathrm{T}$.
\State Train OOD-SDA on $\mathcal{L}$ and predict on $\mathcal{U}$ to get OOD samples, $\mathcal{U'}$ 
\State Consider $m \%$ of the samples from $\mathcal{U'}$ for manual annotation, say $\mathcal{U}'_{m}$
\State Use various labelling strategies to label $\mathcal{U'}$ based on $\mathcal{U}'_m$
\State $\mathcal{L}=\mathcal{L}\cup \mathcal{U}'$ and $\mathcal{U}_{rem}=\mathcal{U}-\mathcal{U'}$
\State Train $\mathcal{M}$ on $\mathcal{L}$ and predict on $\mathrm{T}$ (\textbf{Cycle 1})
\EndProcedure
\end{algorithmic}
\end{algorithm}

\begin{algorithm}[!thb]
\caption{Active Learning based Correction: ALC($\mathcal{L}, \mathcal{U}_{rem}, \mathcal{M}$, $\mathrm{T}$, $\mathcal{TH}$, $\mathcal{MV}$)}
\label{algo:alc}
\begin{algorithmic}[1]
\Procedure{AL Correction}{}
\State Apply Model $\mathcal{M}$ on $\mathcal{U}_{rem}$ to predict intent class label and respective probability score
\For{Each sample in $\mathcal{U}_{rem}$}
\If{score $< \mathcal{TH}$}
\State Apply Classifiers - RF, LR, Bg, KNN and LDA                                
\State Label the sample with Majority Voting predictions (Vote Count $\le$MV)
\State Add Sample and Label to Majority Voting Set ($\mathcal{MVS}$)
\State Add Rejected Samples to $\mathcal{R}$ set
\EndIf
\EndFor
\State Manually Annotate $\mathcal{R}$ set
\State $\mathcal{L} = \mathcal{L} \cup \mathcal{MVS} \cup \mathcal{R}$
\State $\mathcal{U}_{rem} = \mathcal{U}_{rem} - \mathcal{MVS} - \mathcal{R} $
\State Retrain $\mathcal{M}$ on $\mathcal{L}$ and predict on $\mathrm{T}$
\State Repeat Steps 2 to 14 for K times
\EndProcedure
\end{algorithmic}
\end{algorithm}
\vspace{-2mm}

\noindent \textbf{Neural Model} ($\mathcal{M}$): We use the JointBERT \cite{chen2019bert} model as our learning model. The model performs intent classification. We do not use the auxiliary task of slot-filling. 
We use English uncased BERT-Base\footnote{We explore different models but it performs the best} model \cite{devlin2019bert} as the base model for JointBERT. In the case of other languages, we use the bert-base-multilingual-uncased model. For the neural model, we perform a hyperparameter search and report the results of the settings that achieve the best results and then fix the same for all the models. The batch size is 16, the number of epochs is 10, the optimization algorithm used is AdamW and the learning rate is $5e-5$ with cross-entropy as the loss function. 


\subsection{Intent Detection (ID)} \label{sec:ood}

The algorithm (Algorithm \ref{algo:id}: steps 1-8) for our Detection consists of the following Steps: 

\textbf{\underline{Step 1}} \textit{Cycle 0}: With the initial labeled data ($\mathcal{L}$) that consists of examples only from the set of known intents, we train a Neural Model ($\mathcal{M}$). 
    
 \textbf{\underline{Step 2:}} \textit{Out-Of-Domain Sample Detection Algorithm (OOD-SDA)}: In this step, we consider an algorithm to detect Out-Of-Domain samples $\mathcal{U'}$ on the Unlabeled Data ($\mathcal{U}$). 
 
 \textbf{\underline{Step 3:}} \textit{Efficient Labeling of New Intents in OOD}: In this step, we consider $m\%$ of the OOD samples $\mathcal{U'}$ for manual annotation ($\mathcal{U}'_{m}$). The $\mathcal{U'}$ OOD samples can be labeled with new as well as old intents. We experiment with various labeling strategies to add the rest of the OOD samples back to the Cycle 0 training set ($\mathcal{L}$). 
 
\textbf{\underline{Step 4}} \textit{Cycle 1}: We re-train $\mathcal{M}$ on the modified training set and perform prediction on the Test Set. 

\subsection{Active Learning based Correction (ALC)} \label{sec:alc}
\begin{figure}
    \centering
    \vspace{-4mm}
\includegraphics[width=8.4cm, height = 8.5 cm]{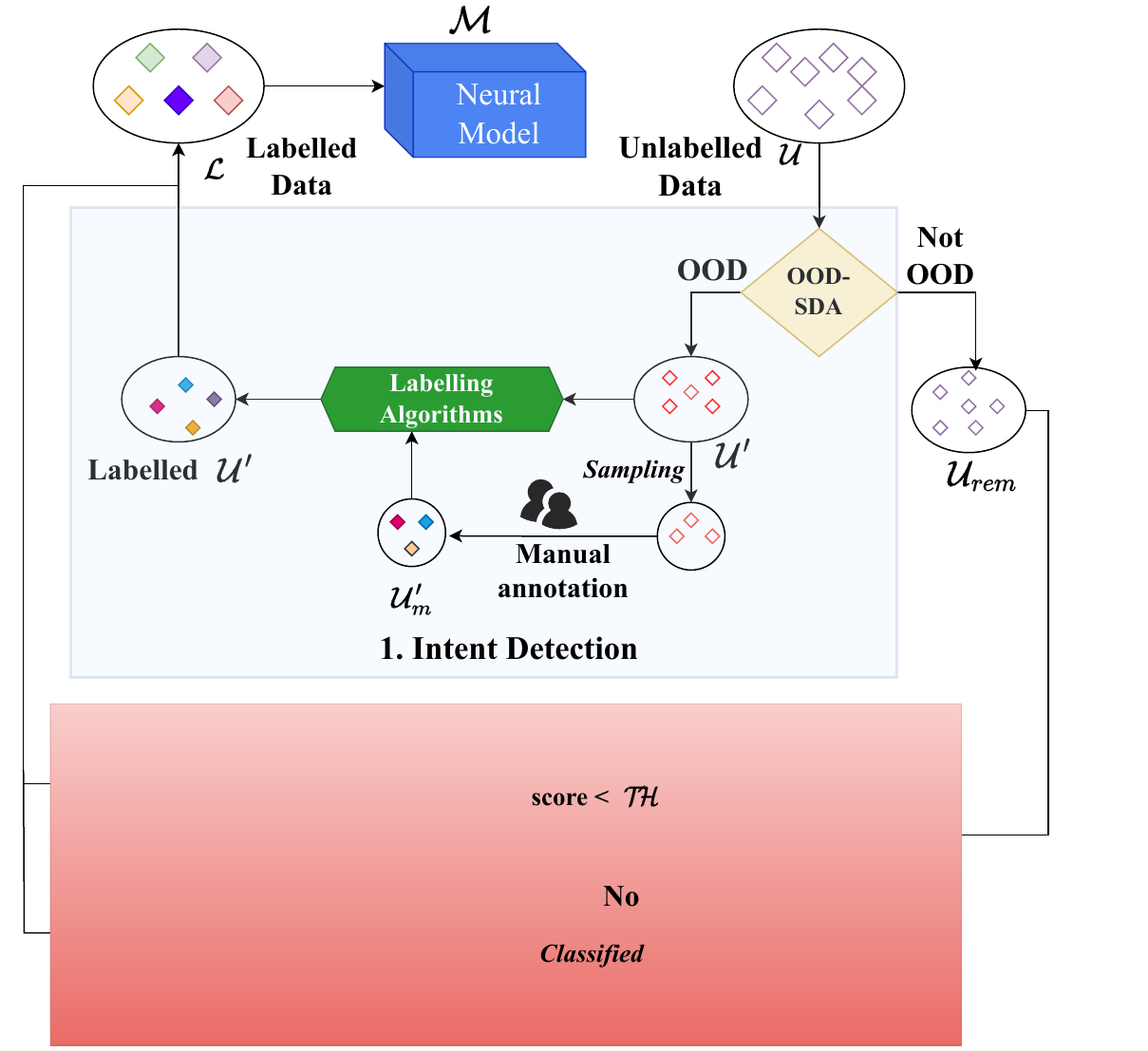}
    \caption{End-to-end framework of IDALC:Intent \\Detection and Active Learning based Correction}
    \label{fig:al_archi_final}
    \vspace{-5mm}
\end{figure}

Next, we devise an active learning-based framework to take care of the rejected intent utterances with low confidence for rectification. As shown in Figure \ref{fig:al_archi_final}, we take the utterances from the unlabeled data and do active learning-based auto corrections of low confidence instances (i.e., the utterances with a score below a threshold after passing through the neural model $\mathcal{M}$) by majority voting. Then we manually annotate the low-confidence and auto-rejected utterances and add both to the training set. 
The steps (1-16) of the Algorithm \ref{algo:alc}) are as follows:
    
\textbf{\underline{Step 1}} \textit{Cycle 1}: We use the retrained Neural Model ($\mathcal{M}$) (Cycle 1 of intent detection) to predict intent scores on the unlabeled dataset after removing OODs ($\mathcal{U}_{rem}$). 

\textbf{\underline{Step 2}} \textit{Cycle 1}: We set a pre-defined threshold ($\mathcal{TH}$), and the samples with score < $\mathcal{TH}$ are then fed to five different classifiers and we use majority voting prediction to identify their labels (\textit{auto-corrected}). Samples that do not find a majority are manually annotated. These auto-corrected and annotated samples are added back to the labeled data and removed from the unlabeled data. 

 \textbf{\underline{Step 3}} \textit{Cycle 2}: We retrain the Neural Model ($\mathcal{M}$) with an updated labelled dataset and redo \textbf{Steps} 1-2 for the same threshold value.
 
\textbf{\underline{Step 4:}} We run this approach for $2\leq K\leq 5$ cycles.   

%% file: 5Experiment.tex
\section{Experimental Results}
We discuss our experimental set-up detailing various algorithms and strategies utilized for both the components, competing baselines, and the experimental results. We categorize the experimental outcomes into two compartments - intent detection and active learning-based correction.

\subsection{Intent Detection}
For identification of new intents, we consider the following settings:

\noindent \textbf{Out-of-Domain Sample Detection Algorithms}  (OOD-SDA): We consider three algorithms for detecting out-of-domain samples. i) Softmax Prediction Probability (MSP) \cite{hendrycks2018baseline}, 
ii) Deep Open Classification (DOC) \cite{shu2017doc} method 
and iii) Local Outlier Factor (LOF) \cite{10.1145/342009.335388}. 
For MSP, we use the minimum prediction score on the training set to decide the threshold and report for the best threshold. We experiment on all English datasets as shown in Table \ref{tab:tab_ood}. DOC performs the best in terms of both accuracy (Acc) and macro-f1(F1), so we use the DOC model in our pipeline and for different experiments.

\begin{table}[!htb]
\centering
\begin{adjustbox}{width=0.8\columnwidth}
\begin{tabular}{|c|c|c|c|c|}
\hline
\textbf{Dataset}& \textbf{Method} & \textbf{Threshold} & \textbf{Acc (\%)} & \textbf{F1 (\%)}         \\ \hline
\multirow{3}{*}{SNIPS} & MSP             & 0.7    & 86.72                         & 89.27  \\ \cline{2-5}
& LOF  & -  & 81.21  & 83.44  \\ \cline{2-5}
& DOC  & -  & \textbf{90.3}                 & \textbf{89.7}                      \\ \hline
\multirow{3}{*}{ATIS} & MSP             & 0.3                                 & 85.74                         & 86.32                              \\ \cline{2-5} 
& LOF                               & -                                   & 78.2                          & 78.05                              \\ \cline{2-5}
& DOC                           & -                                   & \textbf{88.88}                & \textbf{87.19}                     \\ \hline
\multirow{3}{*}{FB} &MSP             & 0.7                                 & 88.63                         & 89.11                              \\ \cline{2-5} 
& LOF     & -     & 81.49  & 82.37  \\ \cline{2-5}
& DOC    & -     & \textbf{93.42}                & \textbf{93.5}                      \\ \hline
\end{tabular}
\end{adjustbox}
\caption{Accuracy(Acc) and Macro-F1(F1) of \\various OOD-SDA on the three datasets}
\vspace{-2mm}
\label{tab:tab_ood}
\end{table}

    \begin{table}[b]
    \centering
    \vspace{-4mm}
\begin{adjustbox}{width=0.9\columnwidth}
\begin{tabular}{|c|c|c|c|c|c|c|}
\hline
\textbf{Class} & \multicolumn{6}{c|}{\textbf{Accuracy (\%)}} \\ \hline
& SNIPS & ATIS & FB-EN & FB-ES & FB-TH & Avg \\\hline
\textbf{RF} &  95.4   &    89.8 & 91.2    & 90.6   & 88.3 & 91.1\\ \hline
\textbf{Bg} &97.9     & 87.8    & 92.3 & 89.5    & 88.4 & 91.2\\ \hline
SVM &   87.4  & 85.8    & 86.42    &    83.5 & 80.4 & 84.7\\ \hline
\textbf{LR} &    97.8 & 96.5    &94.7&    95.9 &  97.4 & 96.5  \\ \hline
XGB &   87.6  & 75.3    & 86.5    & 84.0    & 85.1 & 83.7\\ \hline
NB &    85.1 & 86.9    &83.6&    80.0 & 81.6 &  83.4  \\ \hline
\textbf{LDA} &  97.8   &     97.02& 95.4    &    96.7 &97.6 & 96.9\\ \hline
\textbf{KNN} &   95.1  & 89.6    & 90.5& 82.3    & 92.4 & 90.0  \\ \hline
Gau &  90.2   & 79.1    & 80.2    &     84.0& 85.7 & 83.9\\ \hline
\end{tabular}
\end{adjustbox}
\caption{5-Fold Cross Validation Accuracy over 5 different\\ datasets for various classifiers. The best results are \textbf{bolded}.}
\label{tab:maj-vot-clas}
\end{table}

\begin{table*}[]
\centering
\begin{adjustbox}{width=0.76\linewidth}
\begin{tabular}{|c|c|c|c|c|c|c|c|c|c|c|}
\hline
\multirow{2}{*}{\textbf{Method}} & \multicolumn{2}{c|}{\textbf{FB-EN}} & \multicolumn{2}{c|}{\textbf{FB-ES}} & \multicolumn{2}{c|}{\textbf{FB-TH}} & \multicolumn{2}{c|}{\textbf{ATIS}} &\multicolumn{2}{c|}{\textbf{SNIPS}} \\ \cline{2-11} 
                                 & \textbf{Acc}    & \textbf{F1}   & \textbf{Acc}    & \textbf{F1}   & \textbf{Acc}    & \textbf{F1}   & \textbf{Acc}    & \textbf{F1}
                                   & \textbf{Acc}    & \textbf{F1} 
                                 \\ \hline
SEEN*                           & 84.9            & 86.2              & 75.4            & 79.8              & 71.9            & 77.5              & 84.3            & 71.4 & 85.8 & 86.2            \\ \hline
Zero-Shot-OOD*                        & 86.7            & 87.4              & 62.1            & 70.1              & 68.6            & 72.1              & 84.8            & 81.7 & 87.4 & 88.5             \\ \hline
SENC-MaS* & 82.3 & 83.9 & 69.4 & 70.5 & 64.6 & 72.8 & 80.1 & 69.4 & 82.0 & 81.4\\\hline
SENNE* & 61.4 & 60.5 & 56.1 & 58.9 &58.2 & 59.4 & 53.2 & 51.5 & 52.7 & 55.8 \\\hline
IFSTC  &84.7 & 82.0 &80.4 & 54.3 & 86.1 & 87.2 & 71.1 & 71.4 & 70.4 & 71.1\\ \hline
TARS & 71.2 & 72.3 &61.1 & 60.2 & 81.1 & 80.1 & 78.4 & 72.4 & 79.6 & 81.1 \\ \hline
LLaMA2-FT & 92.2 &78.1 & \textbf{89.1} &71.5  &  83.5 &79.0  & 81.6 & 73.3& 90.6 & 90.1\\ \hline
LLaMA2-FT+ALC(2) & 92.7 & 82.4   & 91.3 & 76.2 & 86.3 & 84.1 & 82.5 & 77.5 & 92.6 & 92.9 \\ \hline
ChatGPT 3.5T-Pr & 70.5 & 69.7 & 70.4 & 64.5 & 51.7 & 52.1 & 68.8 & 51.9 &  70.8& 72.2  \\ \hline
ID (0)                          & 61.9            & 72.4              & 66.6            & 74.2              & 68.1            & 80.2              & 82.6            & 81.1 & 73.7 & 80.2            \\ \hline
ID + RA                         & 66.7            & 52.4              & 56.6            & 52.3              & 48.1            & 60.7              & 52.6            & 51.9 & 73.7 & 78.7            \\ \hline
ID (1) + KM                     & \textbf{95.7}   & \textbf{92.3}     & 87.6            & \textbf{94.9}              & \textbf{89.3}            & \textbf{88.6}     & \textbf{90.3}            & \textbf{86.4} & \textbf{94.7} & \textbf{93.0}             \\ \hline
ID (1) + MV                     & 92.2            & 89.6             & 85.2            & 90.3              & 85.3            & 87.23              & 86.2            & 81.6 & 91.9 & 90.5             \\ \hline
ID (1) + CL                     & 94.1            & 91.8              & 87.3   & 92.1     & \textbf{89.3}   & 88.4              & 89.6   & 83.5 & 94.6 & 92.6   \\ \hline
ID-ALC (2) &  \textbf{98.0} & \textbf{93.1} & \textbf{91.5} & \textbf{96.1} & \textbf{92.5} & \textbf{89.5} & \textbf{92.1} & \textbf{88.9} & \textbf{97.9}& \textbf{95.5} \\\hline

\end{tabular}
\end{adjustbox}
\caption{Results in terms of Overall Accuracy (Acc) and Macro average F1-score (F1) on Facebook - English, Spanish, Thai, ATIS and SNIPS Datasets. (0), (1) and (2) denote the cycle number. All results are in (\%).}
\label{tab:fb_results}
\vspace{-5mm}
\end{table*}

\noindent \textbf{Labelling Strategies}: Once we obtain the OOD samples from the unlabeled data using OOD-SDA, the next step is to add this data back to the training set. The $m\%$ samples chosen uniformly and randomly for manual annotation are set to $20\%$ in our case based on various experiments. The different Labelling Strategies used are discussed below:

    \textbf{K-Means (KM)}: Here, we decide on the number of clusters based on the majority labels of the manually annotated OOD samples. The labels whose frequency is greater than a pre-decided threshold, $t$, are considered the majority labels. The threshold is decided as $t$ = (no of annotated samples) / (total number of labels discovered during annotation). We now run $k$-means algorithm on the entire set of OOD samples on the unlabeled data set and annotate clusters as per their majority labels.
    

    \textbf{Majority Voting (MV)}: With the manually annotated OOD samples, we train several traditional machine learning models to compute the accuracy of the system. We check on a total of twenty different classifiers and the top nine results (accuracy for each dataset and overall average [Avg]) are shown in Table \ref{tab:maj-vot-clas} for Random Forest (RF), Bagging (Bg), Support Vector Machine One-vs-all (SVM), Logistic Regression (LR), XGBoost (XGB), Multinomial NB Classifier (NB), Linear Discriminant Analysis (LDA), K-Nearest Neighbour (KNN), Gaussian (Gau).
    We take the majority vote of the predictions of these classifiers on the rest of the OOD samples to label them. 
    5-fold cross-validation is performed on the labeled data to determine the performance of these classifiers to identify intents. Based on the accuracy scores, the top 5 (out of 9) classifiers (\textbf{Bold} in Table \ref{tab:maj-vot-clas}) are chosen for experiments.
    
    \textbf{Cluster Then Label (CL)}: Here, we start with a pre-defined number of clusters, $k$. We cluster the entire set of OOD samples on the unlabeled data set using $k$-means. We then label those clusters based on their majority labels as per $m\%$ annotated data. If two or more clusters have the same label, we merge these clusters. We set the value of $k$ as double the number of known intents.
    

\subsubsection{\saransh{\textbf{Competing Baselines}}}
\saransh{IDALC aims to enhance task-oriented dialog systems by handling known intents, detecting unseen ones, and adapting in streaming settings with limited annotation. We use various baselines from the last few years - all are closely matching solutions to the problem definition, and compare our framework in different directions - OOD detection, zero/Few-shot approaches, etc. Since all the baselines are state-of-the-art approaches addressing related problems over the last decade, including them in our comparisons is essential to rigorously establish the superiority of our framework. For fair comparison, we evaluated baselines across three categories: (i) streaming/semi-supervised, (ii) zero-shot/OOD detection, and (iii) large language models. These categories capture the main challenges of evolving label spaces, limited data, and annotation efficiency.}

\saransh{\textbf{Notation for Complexity:} We denote $n_{\text{train}}$ and $n_{\text{test}}$ as the number of training and test examples, respectively, with $e$ epochs and sequence length $L$ (fixed at $L \leq 512$). The hidden dimension is $d$ ($d=4096$ for LLaMA2-7B), with $\ell$ transformer layers ($\ell_{\text{BERT}}=12$, $\ell_{\text{LLaMA}}=32$). We cap the number of clusters at $k \leq 20$. For QLoRA, the number of trainable parameters is $p_{ft}$; with rank-$r=4$ LoRA adapters applied to $Q,K,V,O$ projections across layers, this yields $p_{ft} = 8 \cdot d \cdot r \cdot \ell_{\text{LLaMA}} \approx 4.2\text{M}$. ChatGPT-3.5 API calls incur monetary cost $c$. Batch size is kept at 16.}

\saransh{{\textbf{A) Streaming and Semi-Supervised Methods}}\\
\textbf{SEEN} \cite{zhu2020semi} is designed specifically for streaming intent detection, where new labels can appear continuously in the incoming data. It combines two key components: (i) \emph{SEEN-Forest}, an ensemble of random trees that serves as an anomaly detector, and (ii) \emph{SEEN-LP}, a graph-based label propagation module for semi-supervised classification of queries. This hybrid design makes SEEN particularly relevant for dialog systems, as it explicitly addresses the challenge of detecting novel intents with minimal human supervision. In our setup, we initialize their model with a small amount of labeled data in the initial stream (stream 0), incrementally update it with unlabeled data from the next stream (stream 1), and finally evaluate it on the held-out test set. The complexity of label propagation is $\mathcal{O}(n_{\text{train}}^2)$ in dense graphs or $\mathcal{O}(n_{\text{train}})$ when sparsified, while training decision trees costs $\mathcal{O}(n_{\text{train}}\log n_{\text{train}})$.}

\saransh{\textbf{SENC-MaS} \cite{mu2017streaming} builds compact matrix sketches that summarize the distribution of known-class data and then leverages these sketches for both classification and novelty detection. By operating on compressed sketches rather than the raw data, it prioritizes efficiency, making it well-suited for high-throughput dialog streams where retraining large neural models on every update is infeasible. In our experiments, the algorithm incrementally updates its sketches with streaming unlabeled data, after which new queries are either classified or flagged as novel. The update complexity is $\mathcal{O}(n_{\text{train}}\, d)$, reflecting its linear dependence on training data and hidden dimension.}

\saransh{\textbf{SENNE} \cite{cai2019nearest} is a nearest-neighbor ensemble approach tailored for emerging-class detection, especially in scenarios where new intents are closely related to existing ones. This characteristic is important for dialog systems, as subtle distinctions (e.g., \textit{PlayMusic} vs. \textit{AddToPlaylist}) can critically impact performance. SENNE works by constructing ensembles of nearest-neighbor classifiers for each known class and checking whether a query falls within or outside the decision boundary of these ensembles. This makes it particularly strong in handling borderline cases where unseen intents resemble existing categories. The computational cost is $\mathcal{O}(n_{\text{train}}\, d)$ at inference.}

\saransh{\textbf{IFSTC} \cite{xia2021incremental} frames intent detection as a textual entailment problem, which enables few-shot classification of new intents without retraining on the entire dataset. This is especially relevant for IDALC, as both approaches aim to generalize quickly from small amounts of annotated data. IFSTC employs a hybrid strategy: it initializes models with XNLI pretraining and then fine-tunes them using only a few labeled examples of new intents. This combination of cross-lingual pretraining and few-shot adaptation makes it robust across languages and efficient in data-scarce settings. Its fine-tuning cost is $\mathcal{O}(e\, n_{\text{train}}\, L^2\, d\, \ell_{\text{BERT}})$.}

\saransh{{\textbf{B) Zero-Shot and OOD Methods}}\\
\textbf{Zero-Shot-OOD} \cite{tan2019out} is built on an OOD-resistant prototypical network, which learns compact clusters of known-class examples and rejects queries that fall outside them. The method directly benchmarks IDALC’s OOD detection capabilities, since both target the challenge of handling unseen intents with limited supervision. In our experiments, prototypes were constructed using meta-learning episodes, and incoming queries were tested for membership within known clusters. Queries sufficiently distant from all prototypes were flagged as out-of-distribution. Training scales as $\mathcal{O}(e\, n_{\text{train}}\, d)$ and inference as $\mathcal{O}(k\, d)$ per query.
}

\saransh{\textbf{TARS} \cite{halder-etal-2020-task} reformulates text classification as a label-text matching task, where each input query is paired with an intent label and the model decides whether they match. This approach is highly relevant for our evaluation, as it probes whether intent names alone can serve as sufficient supervision. Unlike IDALC, which uses a more elaborate loop of OOD detection, clustering, and annotation, TARS relies directly on semantic alignment between queries and labels. For our benchmarks, we trained TARS models on dialog datasets using BERT-base for English and multilingual BERT for non-English data. The training cost is $\mathcal{O}(e\, n_{\text{train}}\, L^2\, d\, \ell_{\text{BERT}})$, with inference requiring a single forward pass per label.}

\saransh{{\textbf{C) Large Language Model Baselines}}\\
\textbf{LLaMA2-FT}\footnote{https://ai.meta.com/llama/} fine-tunes a LLaMA2-7B model using parameter-efficient methods such as QLoRA and PEFT. While it serves as a strong benchmark in high-resource setups, it is less suited for incremental adaptation in streaming scenarios. Including this baseline highlights the trade-off between raw accuracy and computational cost. To further investigate adaptability, we also report results with IDALC’s ALC module applied on top (LLaMA2-FT+ALC), demonstrating how annotation-driven clustering can complement large pre-trained models. The training complexity is $\mathcal{O}(e\, n_{\text{train}}\, L^2\, d\, \ell_{\text{LLaMA}})$ with $p_{ft}=4.19$M trainable parameters.}

\saransh{\textbf{ChatGPT 3.5 Turbo}\footnote{https://platform.openai.com/docs/models/gpt-3.5-turbo} represents in-context learning with no model retraining. Queries are answered by constructing prompts that include 100–150 training samples along with the test query, following structured instruction templates. Its relevance lies in evaluating the performance of large hosted APIs that incur zero training cost, but at the expense of latency, privacy, and reliance on external infrastructure. The local complexity is negligible, but usage cost scales with $c$, the monetary cost of API calls.}

\saransh{\textbf{D) Random Annotation (RA)} This baseline randomly selects samples from the unlabeled pool for annotation, equal in number to those annotated by IDALC. It provides a natural lower bound for annotation-driven methods by removing any intelligent sampling strategy. While simple, RA allows us to quantify how much of IDALC’s gains are attributable to active annotation selection rather than annotation volume. The additional computational cost apart from BERT classifier training is negligible, involving only uniform random sampling.}

\saransh{\textbf{Evaluation Note: }Some baselines detect only one novel intent at a time. For fairness, we merge all novel intents into a single class and report accuracy and macro-F1 over known vs. merged classes\footnote{This setting favors such baselines, as they do not separate multiple novel intents.}. Unlike IDALC, these methods use no manual annotation; variants with equal annotation performed worse and are not reported. IDALC alternates between neural training, OOD detection, clustering, annotation, and retraining. The dominant cost comes from BERT training, with complexity $\mathcal{O}(e \, n_{\text{train}} \, L^{2} \, d \, \ell_{\text{BERT}})$. The OOD component is another transformer-based classifier, which scales similarly as $\mathcal{O}(e \, n_{\text{train}} \, L^{2} \, d \, \ell_{\text{BERT}})$. For clustering, we apply $k$-means on hidden representations, costing $\mathcal{O}(n_{\text{train}} \, k \, d)$. At inference, optional nearest-neighbor assignment adds $\mathcal{O}(n_{\text{test}} \, d)$. The overall cost is $\mathcal{O}(e \, n_{\text{train}} \, L^{2} \, d \, \ell_{\text{BERT}} + n_{\text{train}} \, k \, d)$.}

\subsubsection{Results and Discussion}
 
 In Table \ref{tab:fb_results}, we show results in multilingual Facebook datasets (English, Spanish, and Thai), ATIS, and SNIPS (average results from Table \ref{tab:snips_results}). In Table \ref{tab:snips_results}, we report the performance of our algorithm on the different types of SNIPS data. We observe that in almost all the datasets, the \textbf{KM} labeling strategy outperforms others in terms of overall accuracy and Macro-F1 score. We report overall accuracy and Macro-F1 as they give a better idea of performance across both frequent and rare classes. We also observe that \textbf{CL} performs significantly well on all the datasets and even outperforms \textbf{KM} in the SNIPS Type 3 setting as in Table \ref{tab:snips_results}. Thus, \textbf{CL} can be a decent alternative to \textbf{KM} and may significantly reduce the overhead of deciding the number of clusters. ChatGPT 3.5 Turbo with Prompt does not produce good results which shows limitations of large language models without finetuning for specific tasks. LLaMA2\footnote{We explore LLaMA2 (LLaMA2-7b) without fine-tuning but results are not good, hence we do not report that.} fine-tuned with a prompting model achieves good performances for a few cases (e.g. - accuracy in Spanish without ALC) but does not produce better results than IDALC for different datasets in Table 4. This may be due to the facts like - intent data skewness, similar intent categories in task-oriented datasets (e.g. - \textit{atis-flight} and \textit{atis-flight-no} in ATIS), and limitation of LLaMA2 base model in case of finetuning with fewer data sample and a quantization parameter. In a few cases, `LLaMA2-FT+ALC(2)' outperforms IDALC like in the case of SNIPS Type 2 (in Table \ref{tab:snips_results}) where new intents are dissimilar to each other and with known intents also. In other scenarios, LLaMA2-FT produces better accuracy than other models (without ALC) in FB-Spanish (Table \ref{tab:fb_results}) but IDALC performs better than `LLaMA2-FT+ALC(2)'. The ID Random annotation (RA) baseline fails to perform well on the datasets since the random selection of samples leads to a uniform distribution of all the known and unknown classes and a lower number of OOD samples are added back to training. 




\begin{table}[]
\centering
\begin{adjustbox}{width=\linewidth}

\begin{tabular}{|c|c|c|c|c|c|c|}
\hline
\multirow{2}{*}{Method} & \multicolumn{2}{c|}{\textbf{Type 1}} & \multicolumn{2}{c|}{\textbf{Type 2}} & \multicolumn{2}{c|}{\textbf{Type 3}} \\ \cline{2-7} 
                        & \textbf{Acc}  & \textbf{F1} & \textbf{Acc}  & \textbf{F1} & \textbf{Acc}  & \textbf{F1} \\ \hline
SEEN*                    & 82.6          & 81.4            & 83.4          & 84.5            & 91.3          & 92.7            \\ \hline
Zero-Shot-OOD*                 & 85.7          & 83.2            & 86.2          & 89.1            & 90.4          & 93.1            \\ \hline
SENC-MaS* & 80.7 & 78.3 & 78.5 & 82.1 & 86.9 & 83.7\\\hline
SENNE* &50.4&51.1&56.1&61.5&51.6& 54.8\\\hline
IFSTC &69.1&7.6&73.6&73.5&68.5& 69.2\\\hline
TARS &78.1&78.3&84.5&84.8&76.2&80.2 \\\hline
LLaMA2-FT & 84.0 & 82.8& 97.1 & 97.1 & 90.6& 90.3\\\hline
LLaMA2-FT+ALC(2) & 86.7&87.2& \textbf{98.9} & \textbf{98.1} &92.2& 93.4\\\hline
ChatGPT 3.5T-Pr &73.3& 76.3&79.2&79.0&59.9& 61.4\\\hline
ID (0)                 & 79.4          & 81.6            & 71.1          & 80.6            & 70.8          & 78.3            \\ \hline
ID (1) + RA & 70.4 & 76.6  & 76.2 & 81.3 & 74.7 & 78.1\\\hline
ID (1) + KM            & 96.3          & 94.8 & 95.3 & 93.2  & 92.4          & 91.1          \\ \hline
ID (1) + MV            & 93.6 & 89.9              & 91.9          & 91.5  & 90.2          & 90.0           \\ \hline
ID (1) + CL            & 93.1          & 91.1              & 94.9          & 92.5  & 95.9 & 94.2
  \\ \hline
ID-ALC (2) & \textbf{98.6}& \textbf{95.9} & 98.8 & 95.7 & \textbf{96.2} & \textbf{94.8}\\\hline

\end{tabular}
\end{adjustbox}
\vspace{1mm}
\caption{Results in terms of Accuracy (Acc) and F1-score \\ (F1) (in \%) on test set for different types of SNIPS Data. \\ (Number) represents Cycle no. \textbf{Type 1} (\emph{unknown intents are \\ similar to known intents}). \textbf{Type 2} (\emph{unknown intents are \\ dissimilar  to each other and also known intents}). \textbf{Type 3} \\ (\emph{unknown  intents  are  similar to each other but dissimilar to \\ the known intents})}
\vspace{-5mm}
\label{tab:snips_results}
\end{table}


Our method `IDALC(KM based)' outperforms other baselines in most of the cases by $\sim$ 5-10\% in terms of accuracy and $\sim$ 4-8\% in terms of macro F1. Also, Table \ref{tab:snips_results} infers that all the baselines perform well for the Type 2 setting in the SNIPS dataset. The reason behind this would be the high level of dissimilarity among the unknown intents and with the known intents which makes it easier to detect these intent classes. Despite the fact that the competing baselines have the added advantage of not having to differentiate between the new intents, IDALC still achieves higher results (in terms of both accuracy and macro-f1) in most of the cases. The improvement of IDALC (KM based) over baselines is statistically significant ($p < 0.05$) as per McNemar’s Test in Table \ref{tab:fb_results} and \ref{tab:snips_results}.

\subsection{Active Learning based Correction}
We use Active Learning Classifier (Majority Voting)
on top of ID (using KM) and LLaMA2-FT to detect intents more accurately as shown in Tables \ref{tab:fb_results} and \ref{tab:snips_results}. It helps to increase the accuracy of known intents and identify new cases. Since IDALC (using KM) performs well for most of the datasets we further evaluate ALC on top of ID. We set different threshold values on top of the Neural Model ($\mathcal{M}$) to identify rejected utterances and optimum results are obtained when the threshold is set to 75\% of the maximum classification probability score for a particular dataset. We pass these rejected utterances through a majority voting classifier (the same five classifiers as used previously). If three or more classifiers detect the same label for a rejected utterance, it is auto-labeled (auto-corrected). The rest of the samples (i.e. Majority voting rejection samples) are annotated manually. These auto-corrected and manually annotated samples are removed from the unlabeled set and added back to the labeled dataset, $\mathcal{L}$ for retraining purposes. Thus the auto-correction criteria, save significant human labeling costs. 

\begin{figure}[t]
	\centering
	\includegraphics[width=0.9\linewidth]{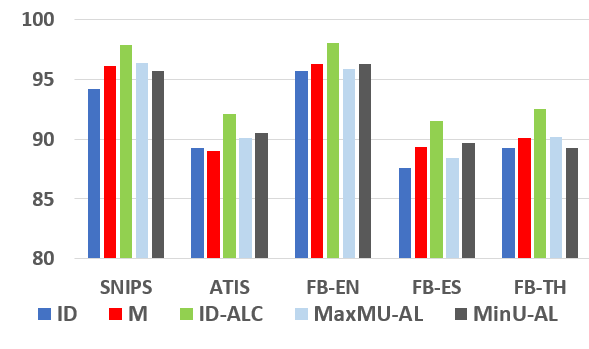}
	\vspace{-3mm}
	\caption{Accuracies (\%) for Final ID, Joint Bert,\\ IDALC, Max and Min Uncertainty-AL models}
	\label{fig:AL-Results1}
	\vspace{-3mm}
\end{figure}

\subsubsection{Baselines}

We apply Maximum Marginal Uncertainty (MaxMU-AL) with minimum interaction \cite{tian2011active} and Minimum Uncertainty (MinU-AL) based Active Learning model \cite{modAL2018} on top of intention detection (ID(1)+KM) output. We use ID(1)+KM, Joint Bert Neural Model ($\mathcal{M}$), MaxMU-AL, and MinU-AL as baselines.

\begin{table}[b]
\centering
\vspace{-2mm}
\begin{adjustbox}{width=0.9\columnwidth}
\begin{tabular}{|c|c|c|c|c|c|}
\hline
\textbf{Dataset} & \textbf{SNIPS} & \textbf{ATIS} & \textbf{FB-EN} & \textbf{FB-ES} & \textbf{FB-TH} \\\hline
Corrections (\%) & 98.21 &  91.86 & 89.71 & 93.91 & 90.62 \\\hline
Acc (\%) & 	95.9 & 81.4 & 92.69 & 94.35 & 82.65\\\hline
\end{tabular}
\end{adjustbox}
\vspace{1mm}
\caption{Majority Voting Corrections of Low Threshold \\Samples and Accuracy (Acc) in \%}
\label{tab:maj-vot-correction}
\end{table}

\subsubsection{Results and Discussion}
\begin{figure}[!hbt]
	\centering
	\includegraphics[width=0.8\linewidth]{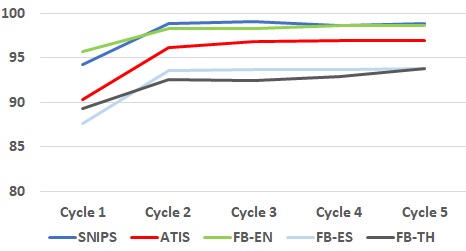}
    \vspace{-1mm}
	\caption{\saransh{Accuracy (\%) across different cycles of \\ ALC module, showing iterative improvement \\ and supporting the choice of limiting the process \\ to two cycles for efficiency.}}
	\label{fig:AL-cycles}
	\vspace{-2mm}
\end{figure}

The addition of the ALC component on top of ID gives a higher boost in performance across all datasets as shown in Tables \ref{tab:fb_results} and \ref{tab:snips_results}. IDALC outperforms other baseline methods (along with different active learning based) across all datasets as shown in Fig \ref{fig:AL-Results1}. We experiment with applying Active learning frameworks (ALC on top of ID) for five different cycles across all datasets as in Fig \ref{fig:AL-cycles}. From the third cycle onwards, improvements are not significant for particular threshold values so we can ignore that to save extra retraining costs. 

\begin{table}[b]
\centering
\begin{adjustbox}{width=0.8\columnwidth}
\begin{tabular}{|c|c|c|c|c|}
\hline
\textbf{Dataset} & \textbf{No MV} & \textbf{MV($\ge$3)} & \textbf{MV($\ge$4)} & \textbf{MV($\ge$5)}  \\ \hline
SNIPS & 46.72  & 1.79 & 4.61 & 7.50 \\ \hline
ATIS &  42.31 & 8.14 & 13.53 & 19.62 \\ \hline
FB-EN & 39.45 & 10.29 & 15.34 & 23.85\\ \hline
FB-ES & 49.68  & 6.09 & 10.78 & 16.75\\ \hline
FB-Th & 41.57  & 10.38 & 16.82 & 22.86\\ \hline
\end{tabular}
\end{adjustbox}
\vspace{2mm}
\caption{Annotation requirement (\%) for different \\conditions of Majority Voting across different datasets}
\label{tab:Annotaion}
\end{table}

We evaluate the ALC model performance by checking what fraction of data is getting automatically labeled by the Majority Voting algorithm and added back to training data. Majority Voting can predict $\sim$90\% low threshold samples with more than 80\% accuracy as shown in Table \ref{tab:maj-vot-correction}. Accuracy is very good for SNIPS (95.9\%), Facebook-English (92.69\%), and Facebook Spanish (94.35\%) where ALC can auto-correct ATIS and FB-Thai with decent accuracy. Thus majority voting module is very important for the overall architecture as it boosts the overall performance without extra neural model retraining costs and optimized annotations. 
Another performance assessment is done for ALC model by evaluating the annotation count for the active learning system. Here, we also annotate majority voting rejected samples during annotation. Table \ref{tab:Annotaion} shows the percentage of unlabeled samples required for annotation when we use different Majority Voting - No Voting (No MV), at least three (MV($\le$3)), at least four (MV($\le$4)) and all classifier agreement (MV($\le$5)) in ALC. The best results are obtained when we use MV($\le$3) - around 10\% of the dataset needs to be annotated to get good results.

\subsection{Human Annotation Cost} 

\saransh{In Table \ref{tab:annot_total}, we report the overall annotation cost of IDALC on the unlabeled data ($\mathcal{U}$). The cost ranges from 6–10\% of the samples, which represents a reasonable trade-off given the substantial performance gains. This indicates that IDALC can reduce labeling effort by more than 90\% compared to a fully supervised setup (i.e. 100\% labeled), where the entire dataset would need to be annotated. Despite this significant reduction in manual labeling, IDALC achieves high accuracy and macro-F1, thereby substantially lowering annotation costs. In contrast, other baselines differ in their labeling requirements: semi-supervised and streaming approaches (e.g., SEEN, SENC-MaS, SENNE) often assume a large initial labeled pool or require frequent labeling during deployment, while zero-shot and few-shot methods rely on minimal labels but typically suffer from lower accuracy and adaptability. To further validate our claim, we ran an experiment where baselines were provided with a random subset of the same size as IDALC’s labeling budget, in addition to the initial labeled set. Their performance was comparable or lower, confirming that IDALC’s advantage stems from its informed sample selection rather than labeling budget size alone.}

\subsection{Interpretability} 
\saransh{Interpretability in intent detection appears in different ways depending on the method. Clustering-based and semi-supervised approaches provide some clarity through cluster centroids or prototype assignments, which highlight typical examples and show how new intents are grouped. Zero-shot and OOD methods rely on embedding similarity, which offers a basic explanation by pointing to the closest known intent or label description, though it does not show token-level details. Large language models mainly provide output-level explanations rather than fine-grained reasoning. In IDALC, interpretability comes from $k$-means clustering, which gives centroid-based insights, and from base classifiers like logistic regression and linear discriminant analysis, where learned weights can be examined. Components such as JointBERT and the ensemble voting in the correction module are primarily designed for improving accuracy. At the user feedback level, IDALC provides additional transparency through the human-in-the-loop setup, where annotators can view errors and guide the system with feedback. While interpretability is valuable and will be explored further, the main focus of this study is on performance: finding new intents without reducing accuracy on known ones, while keeping annotation needs low. This makes the approach practical for real-world applications and adaptable across multiple languages.}

\subsection{Real World Scenarios} 
We explore the applicability of the IDALC approach in various real-world scenarios. 

(i) \saransh{In practical applications, user queries often contain multiple intents. To evaluate our approach in such cases, we used two benchmark datasets for multi-intent detection: Mix-SNIPS~\cite{qin2020agif} and Mix-ATIS~\cite{qin2020agif}. These datasets consist of queries with multiple intent labels. We extended the IDALC architecture to support sentence-level multi-label detection, enabling the system to predict multiple intents for a single query. The modified model achieved 76.45\% accuracy on Mix-SNIPS and 70.37\% on Mix-ATIS, showing its effectiveness and robustness in handling multi-intent queries.}

(ii) \saransh{In scenarios requiring quick responses, our algorithm can detect both known and unknown intents by combining zero-shot or few-shot learning with a majority-voting-based automated method within the IDALC framework. This allows the system to generate responses without extensive manual labeling while maintaining reliable performance. Our approach achieves 81.76\% accuracy on SNIPS, 70.23\% on ATIS, and 78.45\% on Fb-EN. A key advantage is the reduced reliance on manual annotation, which in traditional methods requires labeling 6-10\% of unlabeled data. In large datasets, this translates to hundreds or thousands of samples, creating bottlenecks for real-time applications such as customer service bots, virtual assistants, or emergency response systems. By reducing manual labeling needs, our method improves scalability and responsiveness.}

(iii) \saransh{Our system can also operate in streaming settings where unlabeled data arrives sequentially. At each step, IDALC applies the BERT-based classifier (110M parameters, requiring about 4- 5 GB of GPU memory), making it efficient enough for local or edge devices. High-confidence predictions are accepted directly, while low-confidence cases are checked by an ensemble of lightweight classifiers (Random Forest, Logistic Regression, Bagging, KNN, LDA). If consensus is reached, the corrected intent is returned immediately; otherwise, the sample is flagged as “unknown” or queued for manual annotation. This incremental process supports real-time deployment with low latency, while the BERT model can be updated continuously with new labeled data.  }

\section{Experimental Setup}

\saransh{All experiments were conducted on two Tesla P100 GPUs (16 GB RAM, 6 Gbps, GDDR5) for BERT-based models and one A100 GPU (48 GB RAM, dual 960 GB SSDs) for LLaMA2. Most models trained within 120 GPU minutes, while LLaMA2-7B took about 3–5 hours. LLaMA2-7B was fine-tuned using QLoRA, which applies 4-bit quantization with low-rank adapters (LoRA). A rank of 8 was used for all projection matrices (Q, K, V, O) with LoRA $\alpha = 32$. For in-context prompting, we used ChatGPT-3.5 with default settings (temperature = 1.0, top-$p$ = 0.8, top-$k$ = 50, max output tokens = 10). Code/Data details are in link\footnote{\url{https://github.com/IDALC-TAI/IDALC}}
All fine-tuned models were trained for 10 epochs with AdamW (weight decay = 0.01, $\beta=(0.9, 0.999)$). The learning rate was chosen through a small-scale empirical search and fixed at $5 \times 10^{-5}$, using cross-entropy loss. The dataset was split 80:20 for training and validation, specifically during hyperparameter search. The batch size was fixed at 16 to keep the model lightweight and practical for real-world use. The BERT-based IDALC model already required about 4–5 GB of memory for inference, with additional overhead from optimizer states, activations, and handling long sequences. Larger batch sizes quickly exceeded GPU capacity, while smaller ones only slowed training without any performance gains. This balance ensured the model remained efficient and deployable under realistic resource constraints. Preprocessing used NLTK, SpaCy, and Scikit-learn for tokenization, lowercasing, and removal of stop words, punctuation, and non-alphabetic tokens. Evaluation used Scikit-learn with accuracy, precision, recall, and F1 score.}

%% file: 6Error.tex
\section{Error Analysis}

We perform an error analysis of different parts of our proposed framework to gain more insights.

(i) The trained model, $\mathcal{M}$, sometimes falters when two intents are very similar. On the SNIPS dataset, the sentence ``\textit{i m wondering when i can see hurry sundown}" has gold label ``\textit{SearchScreeningEvent}" but the model predicts as "\textit{SearchCreativeWork}" since both are similar. Similarly, the system wrongly classifies ``\textit{play hell house song}'' as `AddToPlayList' but the actual intent is `PlayMusic' which are very close to each other.

(ii) Majority Voting does not perform well in ID due to less annotated data (novel classes) for training, but works well in ALC as training is done on more data (labeled data ($\mathcal{L}$)). 

(iii) Out-of-Distribution (OOD) detection and clustering approaches exhibit limited effectiveness with ATIS datasets, primarily due to the limited amount of data and the high similarity among known and unknown intents. Our current strategy does not specifically address the challenges posed by datasets with very few samples. We plan to explore further. 


\begin{table}[]
\centering
\begin{adjustbox}{width=0.97\columnwidth}
\begin{tabular}{|c|c|c|c|c|c|c|}
\hline
\multirow{2}{*}{\textbf{Dataset}} & \multirow{2}{*}{\textbf{Type}} & \multirow{2}{*}{\textbf{\# ID}} & \multirow{2}{*}{\textbf{\# ALC}} & \multirow{2}{*}{\textbf{\# Total}} & \multirow{2}{*}{\textbf{\# Unlab}} & \multirow{2}{*}{\textbf{\% annot}} \\
                                  &                                &                                   &                                  &                                    &                                    &                                    \\ \hline
\multirow{3}{*}{SNIPS}            & 1                              & 589                               & 29                               & 618                                & 9110                               & \textbf{6.78}                      \\ \cline{2-7} 
                                  & 2                              & 669                               & 43                               & 712                                & 9110                               & \textbf{7.82}                      \\ \cline{2-7} 
                                  & 3                              & 613                               & 37                               & 650                                & 9110                               & \textbf{7.14}                      \\ \hline
\multirow{3}{*}{FB}               & EN                             & 1600                              & 984                              & 2584                               & 25000                              & \textbf{10.36}                     \\ \cline{2-7} 
                                  & ES                             & 240                               & 55                               & 295                                & 3999                               & \textbf{7.38}                      \\ \cline{2-7} 
                                  & TH                             & 250                               & 54                               & 304                                & 3000                               & \textbf{10.13}                     \\ \hline
ATIS                              & -                              & 100                               & 260                              & 360                                & 3370                               & \textbf{10.70}                     \\ \hline
\end{tabular}
\end{adjustbox}
\vspace{2mm}
\caption{Summary of annotated instances for the IDALC \\ framework across different datasets and configurations.\\ The columns indicate the number of annotated intent (ID) and \\ active learning cycles (ALC), the total annotated samples,\\  the size of the corresponding unlabeled pool, and the \\ percentage of annotated data relative to the unlabeled set.}

\label{tab:annot_total}
\vspace{-6mm}
\end{table}

